\documentclass[12pt]{article}
\usepackage{amsmath}
\usepackage{times}
\usepackage{graphicx}
\usepackage{multirow}
\usepackage[authoryear]{natbib}
\usepackage{rotating}
\usepackage{bbm}
\usepackage{latexsym}
\usepackage[usenames]{xcolor}
\usepackage{amssymb}
\usepackage{amsmath,amstext,amssymb,amsfonts,latexsym,cancel}
\usepackage{longtable}

\newcommand{\captionfonts}{\normalsize}

\makeatletter  
\long\def\@makecaption#1#2{%
  \vskip\abovecaptionskip
  \sbox\@tempboxa{{\captionfonts #1: #2}}%
  \ifdim \wd\@tempboxa >\hsize
    {\captionfonts #1: #2\par}
  \else
    \hbox to\hsize{\hfil\box\@tempboxa\hfil}%
  \fi
  \vskip\belowcaptionskip}
\makeatother   
\newcommand{\figone}[4]{
\begin{figure}[ht]
\begin{center}
\includegraphics[height=#4cm]{#2}
\begin{sl}
\caption{\label{#1}#3}
\end{sl}
\end{center}
\end{figure}
}

\begin{document}
\hspace{13.9cm}1

\ \vspace{10mm}\\

%{\LARGE Predictive Processing in Cognitive Robotics: a Review}
{\Large Predictive Processing in Cognitive Robotics: a Review}
\ \vspace{10mm}\\

\ \\
{\bf \large Alejandra Ciria,$^{\displaystyle 1}$ Guido
  Schillaci$^{\displaystyle 2}$, Giovanni Pezzulo$^{\displaystyle 3}$, Verena V. 
  Hafner$^{\displaystyle 4}$ and Bruno Lara$^{\displaystyle 5}$}\\
{$^{\displaystyle 1}$Facultad de Psicolog\'ia. Universidad Nacional Aut\'onoma de M\'exico.}\\
{$^{\displaystyle 2}$The BioRobotics Institute. Scuola Superiore Sant'Anna. Italy.}\\
{$^{\displaystyle 3}$Institute of Cognitive Sciences and Technologies. National Research Council. Italy.}\\
{$^{\displaystyle 4}$Adaptive Systems Group. Department of Computer Science. Humboldt-Universit\"at zu Berlin. Germany.}\\
{$^{\displaystyle 5}$Laboratorio de Rob\'otica Cognitiva. Centro de Investigaci\'on en
  Ciencias. Universidad Aut\'onoma del Estado de Morelos. Mexico}\\

%\ \\[-2mm]

\thispagestyle{empty}
\markboth{}{NC instructions}
\ \vspace{-0mm}\\
%
%Abstract
\begin{center} {\bf Abstract} \end{center}

 Predictive processing has become an influential framework in cognitive
 sciences. This framework turns the traditional view of
 perception upside down, claiming that the main flow of information processing is realized
 in a top-down hierarchical manner. Furthermore, it aims at unifying
 perception, cognition, and action as a single inferential process. However,
 in the related literature, the predictive processing framework and its
 associated schemes such as predictive coding, active inference, perceptual
 inference, free-energy principle, tend to be used interchangeably. 
 %VVH maybe better: It can be observed that in ...
 In the field of cognitive robotics there is no
 clear-cut distinction on which schemes have been implemented and under which
 assumptions. In this paper, working definitions are set with the main aim of
 analyzing the state of the art in cognitive robotics research working under
 the predictive processing framework as well as some related non-robotic models.
 %VVH restructured the following sentence
 The analysis suggests that, first, both research in cognitive robotics implementations
 and non-robotic models needs to be extended to the study of how
 multiple exteroceptive modalities can be integrated into prediction error
 minimization schemes. Second, a relevant distinction found here is that
 cognitive robotics implementations tend to emphasize the learning of a generative model, while in non-robotics models
 it is almost absent. Third, despite the relevance for active inference, few
 cognitive robotics implementations examine the issues around control and whether it should result from the substitution of inverse models with proprioceptive predictions.
 
 Finally, limited attention has been placed on precision weighting and the
 tracking of prediction error dynamics. These mechanisms should help to 
 explore more complex behaviors and tasks in cognitive robotics research
 under the predictive processing framework.

\section{Introduction}
\label{sec:intro}

Predictive processing has become an influential framework in
the cognitive sciences. A defining characteristic of predictive processing is
that it ``...depicts perception, cognition, and action as the
closely woven product of a single kind of inferential process.''
\citep[p. 522]{clark2018nice}. This idea has caused a profound effect on
the models and theories in different research communities, from neuroscience to
psychology, computational modelling and cognitive robotics.
In the literature, terms such as ``predictive processing'',
``hierarchical predictive processing'', ``active inference'', ``predictive
coding'' and ``free energy principle'' are often used interchangeably. Scholars refer to them either as  \textit{theories} or \textit{frameworks}, occasionally interweaving their core ideas.

In cognitive robotics, a number of architectures
and models have claimed to follow the postulates of these
frameworks.
Research in embodied cognitive robotics focuses on understanding and modeling
perception, cognition, and action in artificial agents.
It is through bodily-interactions with their environment that agents are
expected to learn and then be capable of performing cognitive tasks autonomously
\citep{lara2018embodied,schillaci2016exploration}. 
The aim of this article is to set working definitions and
delimit the main ideas for each of 
these frameworks, so as to be able to analyze the literature of cognitive
robotics and the different implementations in the literature.
This should help to highlight what has been done and what is missing, and
above all, what the real impact of these frameworks in the area of robotics
and artificial intelligence is.
Finally, this manuscript sets the issues and challenges that these new frameworks
bring on the table.

The structure of this paper is as follows. Section \ref{sec:work} sets
the relevant working definitions. In Section \ref{sec:cr} different
models and architectures are analyzed in the light of the above mentioned
frameworks. Section \ref{sec:dis} closes the paper.

\section{Working definitions}
\label{sec:work}
For the purpose of this article,  
\textbf{predictive processing} is considered to be the most general set of
postulates. Predictive processing proposes to
turn the traditional picture of perception
upside down \citep{clark2015embodied}. The standard picture of perceptual
processing is dominated by the bottom-up flow of information which
is transduced from sensory receptors. In this picture of perception, as
information flows upwards, a progressively richer picture of the world is then
constructed from a low-level feature layer processing perceptual input 
to a high-level semantics layer interpreting information
\citep{marr1982vision}.
All together, predictive processing claims to unify perception,
cognition and action under the same explanatory scope
\citep{clark2013whatever,hohwy2013predictive}. 

The predictive processing view of perception states that agents are constantly and
actively predicting sensory stimulation and that only deviations from the
predicted sensory input (prediction errors) are processed 
bottom-up. Prediction error is newsworthy sensory information which provides
corrective feedback on top-down predictions and promotes learning. Therefore,
in this view of 
perception, the core flow of information is top-down and the bottom-up flow of
sensory information is replaced by the upward flow of prediction error.   
The core function of
the brain is minimizing prediction error.
This process has become known as
\textbf{Prediction Error Minimization} (PEM).
In a general sense, PEM has been a scheme used in many machine learning algorithms where the error between the desired output and the output generated by
the network is used for learning (see, for instance, backpropagation algorithms for training neural networks). Different strategies of PEM have been used
in models for perception and action control in
artificial agents (see \citet{schillaci2016exploration} for a review).

Going further, predictive processing suggests that the brain is an active organ
that constantly generates explanations about sensory inputs and then tests
these hypotheses against incoming sensory information
\citep{feldman2010attention} -- in a way that is coherent with
Helmholtz's view of perception as an unconscious form of inference.

\begin{figure}[ht]
\begin{center}
%\fbox{\includegraphics[height=7.5cm]{#2.ps}}
\includegraphics[height=9cm]{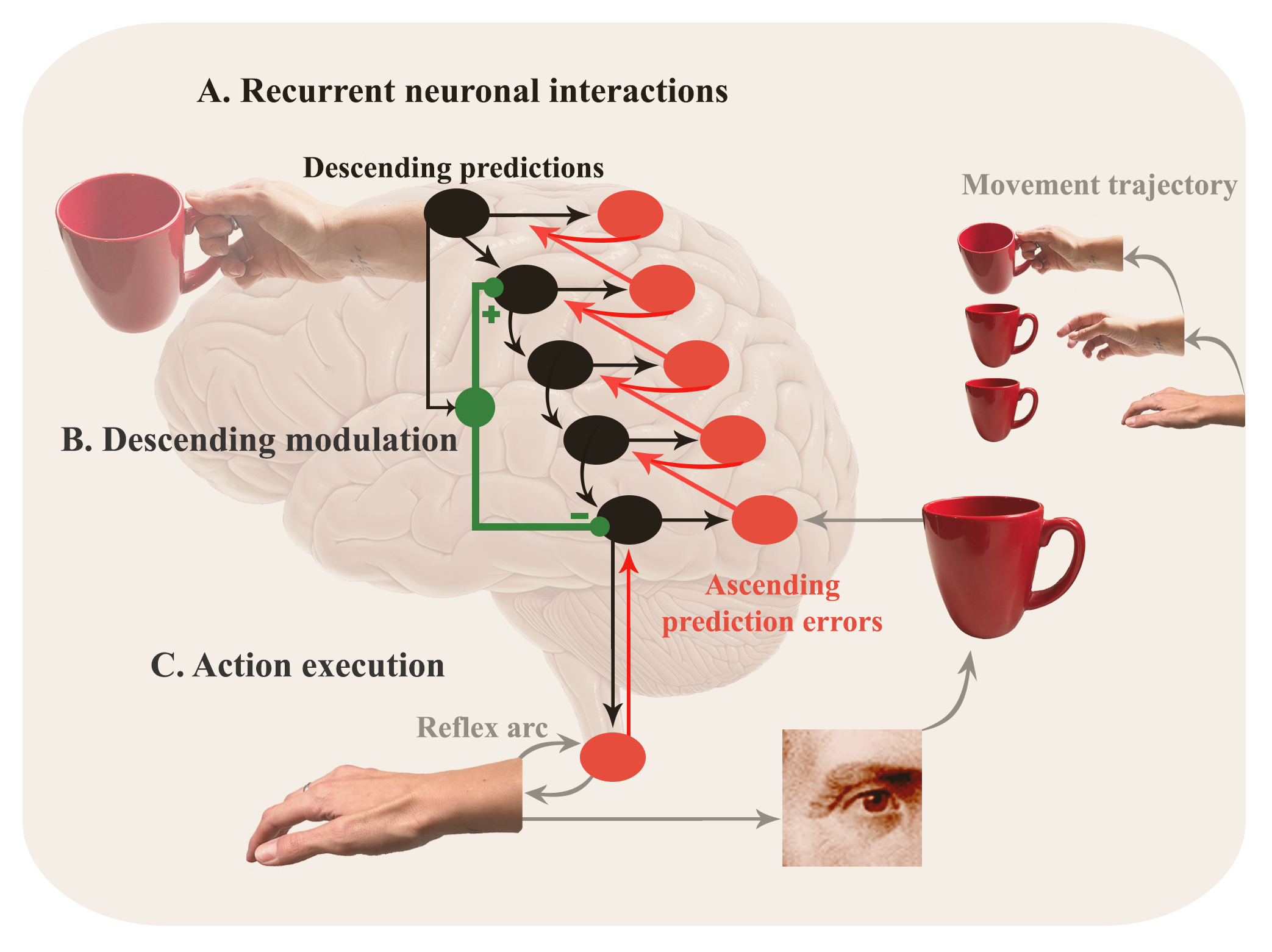}
\caption{Schematic representation of hierarchical neuronal message 
        under the predictive processing postulates.}
\label{fig:actinf}

\end{center}
\end{figure}

Recurrent neuronal interactions with descending predictions and ascending prediction errors following the predictive processing postulates are illustrated 
in a simplified segment of the cortical hierarchy in Figure \ref{fig:actinf}.\textbf{A}. 
Neuronal activity of deep pyramidal cells (represented in black) at higher layers of the cortex encode 
prior beliefs about the expected states of the superficial pyramidal cells (represented in red) at lower layers. 
At each cortical level, prior beliefs encode the more likely neuronal activity at lower levels. 
Superficial pyramidal cells compare descending predictions with the ascending sensory evidence resulting in what is
known as prediction error. The prediction error at superficial pyramidal cells is sent to deep pyramidal cells for 
belief updating (posterior belief). In Figure \ref{fig:actinf}.\textbf{B} descending modulation determines the relative influence of prediction errors 
at lower levels of the hierarchy on deep pyramidal cells encoding predictions. Precision beliefs are encoded by a 
descending neuromodulatory gating or gain control (green) of superficial pyramidal cells. In Bayesian inference, 
beliefs about precision have a great effect on how posterior beliefs are updated. Precision beliefs are considered 
an attentional mechanism which weightens predictions and sensory evidence depending on how certain or useful these 
are for a given task and context. Figure \ref{fig:actinf}.\textbf{C} shows a particular example of active inference for prediction error minimization. 
Perceptual inferences about grasping a cup generate visual, cutaneous, and proprioceptive prediction errors that are 
then minimized by movement. Descending proprioceptive predictions should be fulfilled by being highly weighted to 
incite movement. Then, proprioceptive prediction errors are generated at the level of the spinal cord and minimized 
at the level of peripheral reflexes. At the same time, when the movement trajectory to grasp the cup is performed, 
visual and cutaneous prediction errors are minimized at all levels of the cortical hierarchy. 

Humans and other biological agents have to deal with a world full of sensory uncertainty. 
%Brains have to deal with such an uncertainty to allow agents to efficiently interact with their environment.
In humans, there is psychophysical evidence that shows how Bayesian models can account for perceptual and motor biases by encoding uncertainty in the internal representations of the brain \citep{knill2004bayesian}.

There are several Bayesian approaches centered on the idea that perceptual and cognitive 
processes are supported by internal probabilistic generative models 
\citep{clark2013whatever,clark2015embodied,friston2010freeUnified,hohwy2013predictive,rao1999predictive}.
A generative model is a probabilistic model (joint density), mapping hidden causes 
in the environment with sensory consequences from which samples are generated \citep{friston2010freeUnified}.
It is usually specified in terms of the likelihood probability distribution of observing some sensory 
information given its causes, and a prior probability distribution of the beliefs about the hidden causes of 
sensory information (before sampling new observations) \citep{badcock2017depressed}.
A posterior density is a posterior belief generated by combining the prior and the 
likelihood weighted according to their precision, defined as the inverse variance 
\citep{adams2013computational}. 
A posterior density can be calculated using the Bayes theorem:

\begin{equation}
p(s|O)=\frac{p(O|s)p(s)}{p(O)}
\label{eq:bayes}
\end{equation}

 where $p(s|O)$, also known as the posterior belief, is the probability of hypothesis $s$ with a given evidence or observation 
 $O$. 
 Prior beliefs are updated (thus becoming posterior beliefs) when sensory evidence (likelihood) is available. 
 $p(O|s)$ is the likelihood relating the sensory observation to the hidden causes, this is, the
 probability of the specific evidence $O$. 
 $P(s)$ is the prior distribution of any hypothesis $s$ or prior belief and it can be seen as the prediction of states. 
 $P(O)$ is the probability of encountering this evidence or observation.

This calculation 
is often practically intractable, and variational Bayes is then used for approximately 
calculating the posterior. This method introduces an optimization problem which requires an 
auxiliary probability density termed as the recognition density \citep{buckley2017free}.

Prediction error is the difference between the mean of the prior belief and the mean of the likelihood 
in their respective probability distributions. 
Information gain is measured as the KL divergence between the prior belief and the posterior belief.  
The prior and likelihood distributions have an expected precision, which is encoded as the inverse of their respective 
variance. This precision will bias the posterior belief update. In particular, the posterior belief is updated biased towards the prior belief given its higher expected precision as
compared to the low expected precision on sensory evidence (see Figure \ref{fig:dist}\textbf{A}). 
On the contrary, when the expected precision on prior belief is low and the expected 
precision on sensory evidence is high, the prediction is more uncertain or unreliable, having 
less of an impact on how the posterior belief is updated than the sensory evidence (see Figure \ref{fig:dist}\textbf{B}). 
In both examples in Figure \ref{fig:dist}, although the magnitude of the prediction error is equivalent, the 
information gain is greater in \textbf{B} due to the greater divergence between the prior and the posterior beliefs.

\figone{fig:dist}{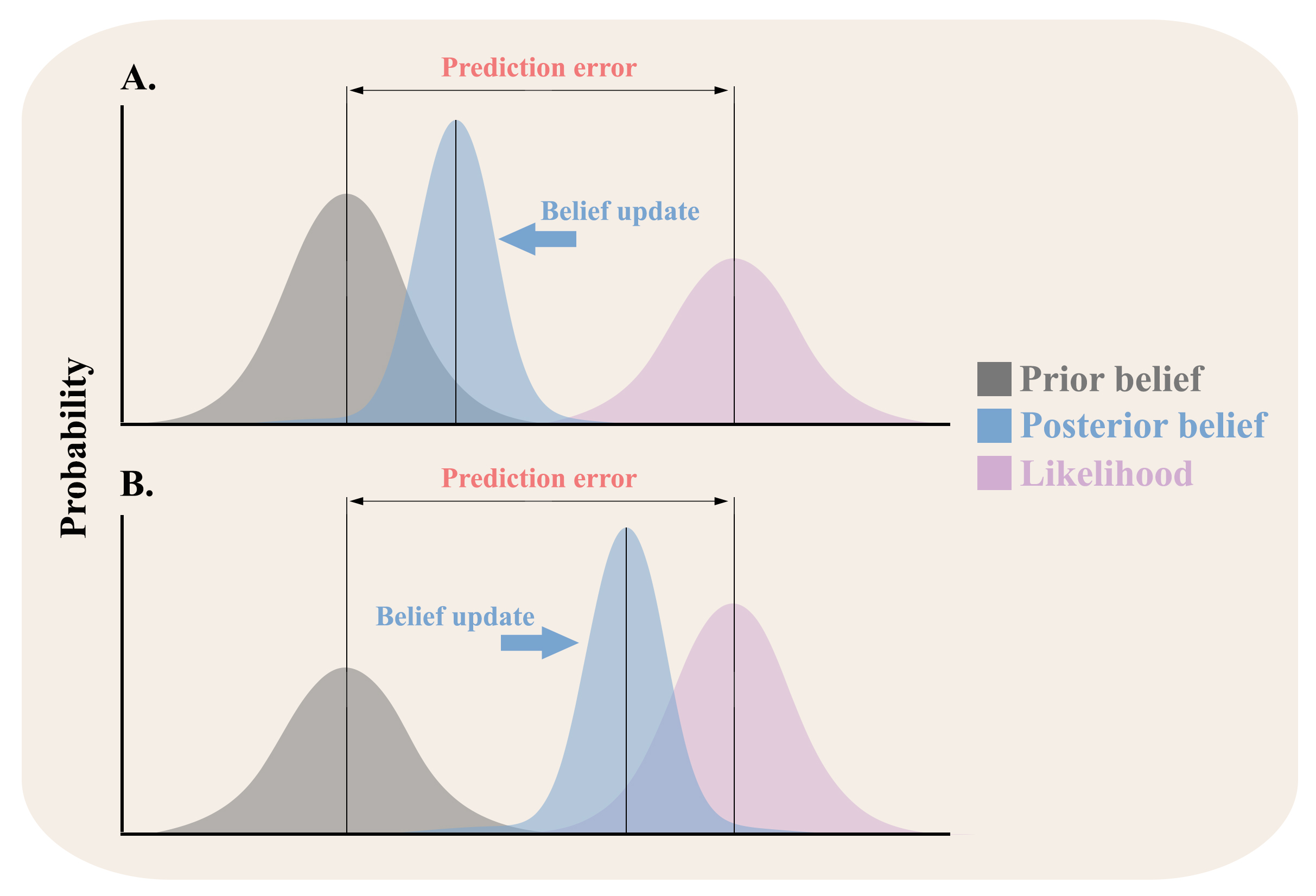}{Relevance of the precision of probability distributions in Bayesian inference. }{7}

In Bayesian inference, there are beliefs 
about beliefs (empirical priors) in terms of having expectations about the beliefs' precision or 
uncertainty \citep{adams2013computational}.
Here,
attention is seen as a selective sampling 
of sensory information, in such a way that predictions about the confidence of
the signals are made to enhance or attenuate prediction errors from different
sensory modalities. In order to attain this sampling, this framework proposes
a mechanism known as \textit{precision weighting}. The information coming from different
modalities are weighted according to the expected confidence given a certain
task in a certain context
\citep{parr2017working,friston2012perceptions,donnarumma2017action}. 

Importantly, precision weights are not only assigned according to their reliability, 
but also by their context-varying usefulness, and are thus considered to be a mechanism for behavior 
control \citep{clark2020beyond}. In the brain, precision weighting might be mediated by a neuromodulatory 
gain control which can be conceived as a Bayes-optimal encoding of precision at a synaptic level 
of neuronal populations encoding prediction errors \citep{friston2014computational}. 
Prediction errors with high precision have a great impact on belief updating, and priors with high 
precision are robust in the face of noisy or irrelevant prediction errors.

Bayesian beliefs are treated as inferences about the posterior probability distribution 
(recognition density) via a process of belief updating \citep{ramstead2020tale}. 
The recognition density is an approximate probability distribution of the causes of sensory 
information which encodes posterior beliefs as a product of inverting the generative model 
\citep{friston2010freeUnified}.
According to the Bayesian brain hypothesis, prior beliefs are encoded as neuronal representations, 
and in light of the new evidence beliefs are updated (posterior density) to produce a 
posterior belief following Bayes’ rule 
\citep{friston2014computational}. 
This means that the brain encodes Bayesian recognition densities within its neural dynamics, 
which can be conceived as inferences of the hidden causes to 
find the best ‘guess’ of the environment \citep{demekas2020investigation}.

According to \citet{friston2006free}, predictive processing must be situated
within the context of the \textbf{free-energy principle}
\citep{williams2018predictive}, given that 'prediction error minimization',
under certain assumptions, corresponds to minimizing free energy
\citep{friston2010free}. 
Predictive processing can be seen as a name for a family of related theories, where the 
free energy principle (FEP) provides a mathematical framework to implement the above ideas.
The free-energy
principle is a biological and a neuroscientific framework in which prediction
error minimization is conceived as a fundamental process of self-organizing
systems to maintain their sensory states within their physiological bounds in
the face of constant environmental changes
\citep{adams2013predictions,friston2009free,friston2010free}. 

Essentially, the free-energy principle is a mathematical
formulation of how biological agents or systems (like brains) resist a natural
tendency to disorder by limiting the repertoire of their physiological and
sensory states that define their phenotypes
\citep{friston2010free}.
In other words, to maintain their structural integrity, the sensory states of any biological
system must have low entropy. Entropy is the negative log-probability of an outcome or 
the average ‘surprise’ of sensory signals under the generative model of the causes of 
the signals \citep{friston2011action}.

Therefore, biological systems are obliged to
minimize their sensory surprise (and implicitly entropy) in order to increase
the probability of remaining within their physiological bounds over long
timescales \citep{friston2009free}.

The main aim of minimizing free energy is
to guarantee that biological systems spend most of their time in their valuable
states, those which they expect to frequent. 
Prior expectations prescribe a
primary repertoire of valuable states with innate value, inherited through
genetic and epigenetic mechanisms \citep{friston2010free}.

Agents are constantly trying to maximize the evidence for the generative model by minimizing surprise. 
The FEP claims that because biological systems cannot minimize surprise directly, they need to minimize an 
upper bound called ‘free energy’ \citep{buckley2017free}. Free energy 
can be expressed  as the Kullback-Leibler divergence between two probability distributions, subtracted by the natural log of the probability
of possible states. As stated in \citet{sajid2020degeneracy}, free energy can always
be written in terms of complexity and accuracy:

\begin{equation}
\begin{split}
F &= D_{ KL}(Q(s)||P(s|o))-  ln P(o)\\
 &= D_{ KL}(Q(s)||P(s)))- E_{Q} [ln P(o|s)]
\end{split}
\label{eq:fep}
\end{equation}

Where $Q(s)$ is the recognition density or approximate posterior distribution, and encodes the prior beliefs an agent 
possesses about the unknown variables. The conditional density
$P(s|o)$ is the probability of some (hidden) state (s) given a certain observation (o), 
and  is refereed to as the generative model. 
The first writing in Eq. \ref{eq:fep} can be read as  evidence bound minus log 
evidence or divergence minus surprise. Rewritten as in the second line it is 
read as complexity, which is the difference between the posterior beliefs and 
prior beliefs before new evidence is available and accuracy, the expected log
likelihood of the sensory outcomes given some posterior about the causes of the 
data \citep{sajid2020degeneracy}.

The recognition density (coded by the internal states) and the generative 
model are necessary to evaluate free energy \citep{friston2010freeUnified}. 
Variational free energy (VFE) provides an upper bound on surprise, and it is formally
equivalent to weighted prediction error \citep{buckley2017free}. VFE is a statistical 
measure of the surprise under a generative model. Negative VFE provides a lower 
bound on model evidence. Minimizing VFE with respect to the recognition density 
will also minimize the Kullback-Leiber divergence between the recognition density 
and the true posterior. Therefore, minimizing VFE makes the recognition density, the 
probabilistic representation of the causes of sensory inputs, 
an approximate of the true posterior \citep{friston2010freeUnified}. 
Optimizing the recognition density makes it a posterior density on the causes of sensory information.
 
Biological agents can minimize free energy by means of two strategies: changing the 
recognition density or actively changing their internal states. Changing 
the recognition density minimizes free energy and thus, reduces the perceptual divergence. 
This is a relevant component of the free energy formulation when expressed as complexity minus accuracy.

Minimizing perceptual divergence increases the complexity of the model, defined as the 
difference between the prior density and the posterior beliefs encoded by the recognition 
density \citep{friston2010freeUnified}. This first strategy is known as \textbf{perceptual inference}, 
this is, when agents change their predictions to match incoming sensory information. 
Given that sensory information can be very noisy and ambiguous, perceptual inferences are 
necessary to make the input coherent and meaningful. 

The second strategy is the standard approach to
action in predictive processing, known as \textbf{active inference}
\citep{adams2013predictions,brown2013active}, which consists of an agent 
changing sensory 
inputs through actions that conform to predictions. This is the same as minimizing
the expected free energy \citep{kruglanski2020all}. 
When acting on the world, free energy is minimized by sampling sensory information that 
is consistent with prior beliefs. An action can be defined as a set of real states that 
change hidden states in the world, which are closely related to control states inferred by
the generative model to explain the consequences of action \citep{friston2012active}. 
Therefore, actions directly affect the accuracy of the generative model, 
defined as the surprise about sensory information expected under 
the recognition density \citep{friston2010freeUnified}.
For survival, valuable actions are those which are expected to provide agents with the capability
to avoid states of surprise.

Every action serves to maximize the evidence of the generative model in such a way that 
policies are selected to minimize complexity. The expected action consequences include 
the expected inaccuracy or ambiguity, and the expected complexity or risk, which are combined 
into the expected free energy \citep{kruglanski2020all}. Thus, expected free energy is 
the value of a policy, describing its pragmatic (instrumental) and epistemic value. 
In other words, actions are valuable if they maximize the utility by exploitation 
(fulfilling preferences), and if they minimize uncertainty by exploration on model 
parameters (information gathering, as in intrinsic motivation strategies) \citep{seth2018being}.
Maximizing epistemic value is associated with selecting actions that increase model 
complexity by changing beliefs, whereas maximizing pragmatic value is associated with actions
that change internal states that align with beliefs \citep{tschantz2020learning}. 
Consequently, the minimization of expected free energy occurs when pragmatic and 
epistemic value are maximized.

Priors are constantly optimized because they are
linked hierarchically and informed by sensory data in such a way that learning
occurs when a system effectively minimizes free energy
\citep{friston2010free}.
Here, motor commands are 
proprioceptive predictions, 
as specific muscle movements (internal frame of reference) are mapped onto an
external frame of reference (e.g. vision). 

Furthermore, it has been suggested that for biological systems ``...it becomes important
not only to track the constantly fluctuating instantaneous errors, but also to
pay attention to the dynamics of error reduction over longer time scales.''
\cite[p. 2856]{kiverstein2019feeling}. Rate of change in prediction error is
relevant for epistemic value and novelty seeking situations. In other words,
this mechanism permits an agent to monitor how good it is in performing an
action, and it has been suggested as the basis for intrinsic motivation and value
related learning \citep{kiverstein2019feeling,kaplan2018planning}. Therefore, prediction error
and its reduction rates might signal the expectations on the learnability of
particular situations \citep{van2017affective}. 

Currently, \textbf{predictive coding} is the most accepted candidate to model  
how predictive processing principles are manifested in the brain, namely those laid out by
the FEP \citep{friston2009free,buckley2017free}.
It is a framework for understanding redundancy
reduction and efficient coding in the brain \citep{huang2011predictive}
by means of neuronal message passing among different levels of cortical hierarchies
\citep{rao1999predictive}.
'Hierarchical predictive coding' suggests that the brain
predicts its sensory inputs on the basis of how higher-levels provide
predictions about lower-levels activation until eventually making predictions
about incoming sensory information
\citep{friston2002functional,friston2005theory}.
Active inference enables predictive coding in a prospective way, where actions  
attempt to fulfill sensory predictions by minimizing prediction 
error \citep{friston2011action}.

 In this framework,
the minimization of prediction error occurs through recurrent message passing
within the hierarchical inference \citep{friston2010free}. Therefore, the changes in
higher-levels are driven by the forward flow of the resultant prediction
errors in the lower-level to optimize top-down predictions until the
prediction error is minimized
\citep{friston2002functional,friston2010free}.

Predictive coding is closely related to Bayes formulations,
from the explanation of how ``hierarchical probabilistic generative models''
are encoded in the brain to the manner in which the whole system deals with
uncertainty. 
Furthermore, the PEM hypothesis suggests that the brain can be conceived as being
``literally Bayesian'' \citep[p. 17]{hohwy2013predictive}.

However, there is an increasing number of predictive coding variants, for example,
there are differences in the algorithms and in the type of generative model
they use \citep{spratling2017review}, and in the excitatory or inhibitory
properties of the hierarchical connections
(e.g. \cite{rao1999predictive,spratling2008predictive},
among others). ``These issues matter when it comes to finding definitive
empirical evidence for the computational architectures entailed by predictive
coding'' \citep[p. 3]{friston2019waves}. 

All of these \textit{frameworks} provide new ways to solve the
perception-action control problem in cognitive
robotics \citep{schillaci2016exploration}.  In the last couple of
decades, the standard solution was the use of paired inverse-forward models
in what is known as Optimal Control Theory (OCT). In OCT, a copy of a motor
command predicted by an inverse model or controller is passed to a forward
model that in turn, predicts the sensory consequences of the execution of the
movement \citep{wolpert95,Wolpert98multiplepaired,Kawato99}. This leads to
multiple implementations using artificial agents with different
computational approaches
\citep{Demiris2006361,Moller08Bootstrapping,escobar2016self,schillaci2016bodyrepresentations}.
OCT presents a number of difficult issues to solve, such as 
the ill-posed problem of learning an inverse model. 

On the other hand, in
predictive processing,
optimal movements are understood in terms of inference and
beliefs, and not by the optimization of a value function of states as being the
causal explanation of movement \citep{friston2011optimal}. 
Therefore, there are no
desired consequences, because experience-dependent learning generates
prior expectations, which guide perceptual and active inference \citep{friston2011action}.
Contrary to OCT, in predictive processing there are no rewards or cost functions to optimize behavior. 
Optimal behavior minimizes variational free energy, and cost functions are replaced by 
priors about sensory states and their transitions \citep{friston2012active}.
Understanding movement as a matter of beliefs for generating inferences removes the problem of learning 
an inverse model.

Therefore, predictive processing suggests that there is
no need for an inverse model and, thus, for any efference copy of the motor
command as input to a forward model. The mere existence of the 
efference copy of the motor command is nowadays a  controversial issue \citep{dogge2019moving,pickering2014getting}.
The core mechanism in predictive processing is an Integral Forward
Model \citep{pickering2014getting}, or better known as a generative model, in
which motor commands are replaced by proprioceptive top-down predictions, mapping prior
beliefs to sensory consequences
\citep{friston2011optimal,clark2015embodied,friston2012active}.
Top-down predictions can be seen as control states based on an
extrinsic frame of reference (world-centered-limb position) that are
translated into intrinsic muscle-based coordinates which are then fulfilled by
the classical reflex arcs \citep{friston2011optimal}.  Minimizing proprioceptive prediction error
brings the action about, which is enslaved to fulfill sensory predictions \citep{friston2011action}. 

%VVH reviewed text until here (Oct 18)

\section{Implementations}
\label{sec:cr}

In this section, we review implementation studies inspired on the models and
frameworks described in the previous section.  
Different review papers can be found in the literature. 
This work focuses mostly on robotics research, which has been developing quite rapidly in the
last couple of years. We review also a  
number of non-robotic studies, in particular those having important aspects
that have not received enough exploration in robotics.  
By highlighting them, this work aims at encouraging new experimental research
in embodied cognitive robotics. 

We are certain that there could be work which is not mentioned in this
article. The omission is not intentional. Articles have been selected under
two criteria. First, the authors mention in their work any of the frameworks
described in the previous section. Second, although the authors do not
explicitly mention these frameworks, it is our understanding that these works
could well enter the discussion and bring interesting topics and questions to
the table. This includes also some non-robotic works. 
Deriving from the descriptions in the previous section, the following items
have been considered as relevant to analyze the literature  in cognitive
robotics:

\begin{itemize}

\item {\bf (Bay)} Bayesian/Probabilistic framework. Does the study adopt a Bayesian
  or probabilistic formalization? 

\item {\bf (PW)} Precision weights. Top down predictions and bottom-up prediction errors
  are dynamically weighted  according to their expected reliability.

\item {\bf(FofI)} Flow of information. Predictions flow top-down while the
  difference between predictions and real sensory information -- i.e., prediction
  error -- flows bottom-up in the model. 

\item {\bf (HP)} Hierarchical processing. The model presents a hierarchical
  structure for the processing of information. 

\item {\bf (IM)} Inverse model. The work discusses the benefits or challenges of 
 using an inverse model, as it is the case in OCT.

\item {\bf (Mod)} Modalities. Which modalities are tackled in the proposed model.

\item {\bf (BC)} Beyond motor control and estimation of body states. Most of the
  reviewed studies adopts predictive processing frameworks to control robot
  movements. This attribute is defined to highlight those studies that make a step
  further by addressing aspects of the framework that may help understanding
  or implementing higher-level cognitive capabilities. 
  
\end{itemize}

The selected studies are summarized in Tables \ref{tab:pp_checkmarks} and \ref{tab:training}. 
In particular, Table \ref{tab:pp_checkmarks} classifies each study according to the attributes mentioned above. Table \ref{tab:training} provides an overview of some implementation details of these works: 

\begin{itemize}
    \item Training: the generative model used in the study is either pre-coded or trained. If applicable, this specifies what  type of learning algorithm (i.e., online or off-line) has been employed;
    \item Data generation: if applicable, this specifies how the training data has been generated;
    \item Agent: what type of artificial system has been used in the experiment;
    \item Generative model: the name, or acronym, of the generative model that has been implemented in 
    the study. Some studies may have not implemented any generative model, but
    used instead the forward kinematics provided by the robot manufacturer. 
    \item Aim: what cognitive or motor task has been modelled.
\end{itemize}

\begin{table}[p]
\scriptsize
\begin{tabular}
{ |p{4cm}|p{0.5cm} |p{0.5cm} |p{0.6cm} |p{0.4cm}
    |p{0.4cm} |p{0.85cm}| p{0.4cm}| p{3cm}|}  
  \hline \textbf{Article} &
     \textbf{Bay}&\textbf{PW}&\textbf{FofI}& \textbf{HP}&
     \textbf{IM}&\textbf{Mod} & \textbf{BC} & \textbf{Aim}\\\hline 
     \hline
     \textbf{Robotic studies} 
     \\\hline
     \hline
     \citet{tani1999learning}&-&-&-&\checkmark&-&V&-&Safe navigation \\\hline
     \citet{ahmadi2019novel}& \checkmark &-&\checkmark &\checkmark &-&PV& - &
     Movement imitation
     \\\hline
     \citet{ahmadi2017can}& - &-&\checkmark &\checkmark &-&
     PV & - & Movement imitation \\\hline
     \citet{baltieri2017active}  & \checkmark&\checkmark &
     \checkmark& -& \checkmark & L& - & Gradient following  \\\hline 
     \citet{hwang2018dealing}& - &-&\checkmark &\checkmark &-& PV & - & Gesture imitation \\\hline
     \citet{idei2018neurorobotics} & \checkmark & \checkmark & \checkmark & - & - & PV & \checkmark & Simul. of autistic behav.
     \\\hline
     \citet{lanillos2018adaptive} & \checkmark & - & \checkmark & - & - & PV(T) & - & Body pose estimation
     \\\hline
     \citet{lanillos2020robot} & \checkmark & - & \checkmark & - & - & PV & \checkmark & Self-other distinction
     \\\hline
     \citet{murata2015predictive} & \checkmark &-&\checkmark &\checkmark &-& PV & - & Human-robot interact. \\\hline
     \citet{ohata2020investigation} & \checkmark & - & \checkmark  & \checkmark & \checkmark  & PV & \textcolor{gray}{\checkmark} & Multimodal imitation \\\hline
     \citet{oliver2019active} & \checkmark & - & \checkmark & - & - & PV & - & Visuo-motor coordin.
     \\\hline
     \citet{park2018learning} & - &-&\checkmark &\checkmark &-& PV & - & Arm control \\\hline
     \citet{pezzato2020novel} & \checkmark & - & \checkmark & - & \checkmark & P & - & Arm control \\\hline
     \citet{pio2016active}  & \checkmark & \checkmark & \checkmark & \checkmark & \checkmark & PV & - & Control and body estim.
     \\\hline
     \citet{sancaktar2019end} & \checkmark & - &\checkmark& - & - & PV & - & Control and body estim.
     \\\hline
          \citet{schillaci2020} & - & - & - & - & \checkmark & PV & \checkmark & Goal regulation,emotion\\\hline
          \cite{annabi2020autonomous} & \checkmark & - & - & - & \checkmark & PV & - & Simul. arm control\\\hline
          
     \citet{zhong2018afa}  & - & - & \checkmark & \checkmark & - & PV & - & Movement generation\\\hline
     \hline
     \textbf{Non robotic studies}
     \\\hline
     \hline
     
     \citet{allen2019body}  & \checkmark & \checkmark & \checkmark & - & - & IV & \checkmark & Emotional inference\\\hline
     \citet{baltieri2019active}  & \checkmark & - & \checkmark &  - & \checkmark& P & - & 1 DoF Control \\\hline
      \citet{friston2015active}  & \checkmark & \checkmark & \checkmark & \checkmark & - & RO & \checkmark & Explorat. vs exploitat. \\\hline
     \citet{huang2011predictive}& \checkmark & -&\checkmark &\checkmark &-&V & - & Visual perception\\\hline
     \citet{oliva2019development} &  \checkmark  &  \checkmark & -  & - & - & V& \checkmark & PW development \\\hline
     \citet{philippsen2019predictive} & \checkmark  &  \checkmark &  - & - & - & V & \checkmark &  PW \& represent.drawing \\\hline
     %\citet{remmelzwaal2019ctnn} & & & & & & & & \\\hline
     \citet{tschantz2020learning}  & \checkmark & - & \checkmark & - & - & RO & \checkmark& Epistemic behaviours
     \\\hline
   %\end{tabular}
\end{tabular}
\caption{\scriptsize \textit{Legend}. \textbf{Bay}: Bayesian/probabilistic framework; \textbf{PW}: implements precision-weighting; \textbf{FofI}: tackles bottom-up/top-down
    flows of information; \textbf{HP}: implements hierarchical processing; \textbf{IM}:
    discusses about the need of inverse models; \textbf{Mod}: modalities addressed in the experiment (P: proprioception, V: visual; T: tactile; I: interoceptive; L: luminance as chemo-trail; RO: simulated rewards and observation); \textbf{BC}: the study goes beyond motor control and estimation of body states.}
    \label{tab:pp_checkmarks}
\end{table}

\begin{table}
\scriptsize
\begin{tabular}
{|p{3.4cm}|p{1.2cm}|p{2.8cm}|p{1.7cm}|p{3.4cm}|} 
     \hline
     \textbf{Article} & \textbf{Train} & \textbf{Data generation} & \textbf{Agent} & \textbf{Generative model}  \\\hline
     \hline
     \textbf{Robotic studies}
     \\\hline
     \hline
     \citet{tani1999learning}&On-line&Direct learning&Mobile.ag.&RNN\\\hline
     \citet{ahmadi2019novel}&Off-line & Direct teaching & Humanoid & PV-RNN  \\\hline
     \citet{ahmadi2017can} &Off-line & Direct teaching & Humanoid & MTRNN  \\\hline
     
     \citet{baltieri2017active} & On-line& Exploration & Mobile.ag.& Agent dynamics \\\hline
     \citet{hwang2018dealing}&Off-line & Direct teaching &Simul.hum.& VMDNN \\\hline
     \citet{idei2018neurorobotics} & Off-line & Recorded sequences & Humanoid & S-CTRNN with PB 
     \\\hline
     \citet{lanillos2018adaptive} & Off-line & Random movements  & Humanoid  & Gaussian Process Regress.  \\\hline
     \citet{lanillos2020robot} &  Re-train
     & left-right arm mov. & Humanoid & Mixt. Dens. Net.,DL class.  \\\hline
     \citet{murata2015predictive}&Off-line &Motionese & Humanoid & S-CTRNN \\\hline
     \citet{ohata2020investigation} & Off-line & Human demonstrations & Humanoid & Multiple PV-RNN  \\\hline
     \citet{oliver2019active}
     & None & N.A. & Humanoid  & Forward kinematics  \\\hline
     \citet{park2018learning}&Dev.learn. & Sets of actions & Humanoid & RNNPB  \\\hline
     \citet{pezzato2020novel} & None & N.A. & Industr.rob. & Set-points \\\hline
     \citet{pio2016active} & None & N.A. &  Humanoid & Forward kinematics
     \\\hline
     \citet{sancaktar2019end} & Off-line & Rand.expl.,direct teach. & Humanoid  & Convolutional decoder  \\\hline
     \citet{schillaci2020} & On-line & Goal-directed expl. & Simul.robot & Conv.AE,SOM, DeepNN \\\hline
     \citet{annabi2020autonomous} & Off-line & Exploration & Simul.arm & SOM, RNN \\\hline
     \citet{zhong2018afa} & Off-line & Recorded sequences & Simul.robot & Convolutional LSTM \\\hline
     \hline
     \textbf{Non robotic studies} \\\hline \hline
     \citet{allen2019body} & None & N.A. & Minim.agent & Markov Decision Process\\\hline
      \citet{baltieri2019active}      & On-line& N.A. & 1 DoF agent& System dynamics      \\\hline
      \citet{friston2015active} & None & N.A. & Simul.rat & POMDP \\\hline
      
     \citet{huang2011predictive} & Off-line & Image dataset &  - & Hierarchical neural model \\\hline
     
     \citet{oliva2019development}& Off-line & Pre-coded trajectories& Sim.drawing &  S-CTRNN \\\hline
     \citet{philippsen2019predictive}& Off-line & Human demonstrations & Sim.drawing & S-CTRNN \\\hline
     %\citet{remmelzwaal2019ctnn} & & & & \\\hline
      \citet{tschantz2020learning} & On-line & RL exploration &OpenAIsim & Gaussian,Laplace approx.
      \\\hline
   \end{tabular}
   \caption{\scriptsize Legend. \textbf{Training}: which type of training -- if applicable
    -- has been performed on the generative model; \textbf{Data Generation:}
     how the training data has been generated; \textbf{Agent}: which type of
     artificial system has been used; \textbf{Generative model}: the name of
     the machine learning tool -- if applicable -- that has been adopted for
     training the generative model; \textbf{Aim}: which cognitive or motor task
     has been modelled. \textit{N.A.}: not applicable.} 
      \label{tab:training}
%\end{table*}
%\end{sidewaystable*}
\end{table}

\subsection{Robotic implementations}

The analysis of the literature starts with one of the first robotic implementations
of predictive processing. \citet{tani1999learning} present a two-layers hierarchical 
architecture that self-organizes expert modules. Each expert module is a Recurrent Neural Network (RNN).
The bottom layer of RNNs is trained and responds to different types of
sensory and motor inputs. The upper set of experts serves as a gating mechanism for the lower level RNNs. The computational model has been deployed onto a simulated mobile robot 
for a navigation task.% navigates in two different rooms. 
The architecture is trained in an on-line fashion. After a short period of time, the gating experts specialize in navigating through corridors, right and left 
turns and T-junctions. The free parameters of the architecture are trained on-line using the back-propagation through time algorithm
\citep{rumelhart1986learning}. 
%For on-line learning, the authors use a posterior probability that a RNN generated a -predicted- target vector. 
However, as the authors point out, a limitation of the architecture is that it only uses the 
bottom-up flow of information, without integrating top-down predictions to modulate the activation of lower levels. \citet{tani2019accounting} provides a thorough review of related neurorobotics experiments, many of which carried out in the authors' laboratory. A very interesting implementation is described in
\citet{hwang2018dealing}, which the authors refer to as a predictive coding model. The adopted network is a multi-layer hierarchical architecture encoding visual and proprioceptive information. Although the work is far from the formulations laid in the free-energy principle \citep{friston2009free}, the VMDNN (Predictive Visuo-Motor Deep Dynamic Neural Network)
performs very similar operations. 
These include the generation of actions following a prediction
error minimization scheme and the usage of the same model structure for action generation and
recognition. Authors claim that
``the proposed model provides an online prediction error minimization mechanism
by which the intention behind the observed visuo-proprioceptive patterns can
be inferred by updating the neurons' internal states in the direction of
minimizing the prediction error'' \citep[pp. 3]{hwang2018dealing}. It is worth noting that such an update does not refer to model weights but only to the state of the
neurons. 
The training of the model is performed in a supervised fashion. The error being minimized is the difference between a
signal generated through kinesthetic teaching (i.e., where a human experimenter manually directs the movements of the robot limb) and the model predictions.
A very interesting aspect of the network are the lateral connections between
modalities at each layer of the hierarchy.

Another relevant work from the same group 
\citep{ahmadi2019novel} stands out for its formulation of active inference and a training strategy based on variational Bayes Recurrent Neural Networks.

Finally, \citet{ahmadi2017can} propose a multiple timescale
recurrent neural network (MTRNN) which consists of multiple levels of
sub-networks with specific temporal constraints on each layer. The model
processes data from three different modalities and is capable of
generating long-term predictions in both open-loop and closed-loop fashions. During closed-loop output generation, internal states of the network can be inferred through error regression. 
The network is trained in an open loop manner, modifying free parameters using
the error between desired states and real activation values.

A common characteristic of the implementations reviewed so far is
that learning and testing are decoupled. During the testing phase,
prediction errors flow bottom-up and the network's 
``internal state is modified in the direction
of minimizing prediction error via error regression''
\citep[pp. 4]{ahmadi2017can}. This implies that  network's weights are not modified
after training.
In most of their works, Tani and colleagues use mathematical formulations based on 
connectionist networks and different from those proposed by \citet{friston2009free}; 
nonetheless, the  work is conceptually very related to predictive coding and active inference. In more 
recent works (e.g. \citep{matsumoto2020goal,jung2019goal}),
authors use explicitly variational inference. 
An illustrative architecture, that comprises most of the characteristics of the networks used by 
these authors can be seen in Figure 1 in \citet{hwang2018dealing}.

A similar approach is presented by \citet{murata2015predictive}, who propose a RNN-based model named
stochastic continuous-time RNN (S-CTRNN). The framework integrates
probabilistic Bayesian schemes in a recurrent neural network. Networks training
is performed off-line using temporal sequences under two learning conditions, i.e., with and
without presenting actions that reveal distinctive characteristics amplifying or exaggerating
meaning and structure within bodily motions (also named \textit{motionese}
\citep{brand2002evidence}). Training data is obtained through kinesthetic
teaching on the robot directed by an experimenter.
The loss function of the optimization process considers the sum of log-uncertainty and
precision-weighted prediction error. This is
formally equivalent to free energy as proposed in active inference. 

In trying to explain the underlying mechanisms causing different types of behavioral rigidity of the autism spectrum, \citet{idei2018neurorobotics} adopt a S-CTRNN with parametric bias (PB) as the computational model
for simulating aberrant sensory precision in a humanoid robot. In this study, S-CTRNN learn to estimate sensory variance (precision) and to adapt to different environments using prediction error minimization schemes. Learning is performed in an off-line fashion using pre-recorded perceptual sequences. "The objective of the learning is to find the optimal values of the parameters (synaptic weights, biases, and internal states of PB units) minimizing negative log-likelihood, or precision weighted prediction error". Once trained, the network is capable of reproducing target visuo-proprioceptive sequences.
In the test phase following the learning one, only the internal states of the PB units are updated in an online fashion, while keeping the other parameters as fixed. The study simulates increased and decreased sensory precision by altering estimated sensory
variance (inverse of their precision). This is performed by modulating a constant in the activation function of the variance units of the trained model. Interestingly, the authors report abnormal behaviors in the robot, such as freezing and inappropriate
repetitive behaviors, correlated to specific modulation of the sensory variance. In particular, increased sensory variance reduces the precision of prediction error, thus freezing the PB states of the network and, consequently, the robot behavior. Decreasing sensory variance, instead, leads to unlearned repetitive behavior, likely due to the fixation of the PB states on sub-optimal local solution during prediction error minimization.

\cite{ohata2020investigation} extends the Predictive coding-inspired Variational Recurrent Neural Network (PV-RNN) presented by \cite{ahmadi2019novel} 
in a multimodal imitative interaction experiment with a humanoid robot.%, where visuo-proprioceptive sensory inputs are fed into the framework implementing predictive coding and active inference.
Modalities (proprioception and vision) -- each encoded with a
multi-layered PV-RNN -- are connected through an associative PV-RNN module. The associative module generates the top-down prior, which is then fed to both the proprioception and vision modules. Each sensory module also generates top-down priors conditioned by the other flows.  Authors show how meta-priors assigned to the proprioception and vision modules impact the learning process and the performance of the error regression. Modulating the Kullback-Leibler divergence (KLD) term in the error minimization scheme leads to a better regulation of multimodal perception, which would be otherwise biased towards a single modality. %In fact, visual and proprioceptive modalities are fundamentally different in respect to their intrinsic randomness, as visual inputs fluctuate more \citep{ohata2020investigation}. 
Stronger regulation of the KLD term also lead to higher adaptivity in a human-robot imitation experiment.

\citet{park2012predictive} proposes an
architecture based on self-organizing maps and transition matrices for studying three different capabilities and phenomena,
i.e., performing trajectories, object permanence and imitation. Interestingly, the
architecture features a hierarchical self-organized representation of state
spaces. However, no bidirectional (top-down/bottom-up) flow of information as in the previous studies is implemented. Moreover, the models are in part pre-coded.% not learned as in the case of provided action sequences.
In a more recent study,  \citet{park2018learning} adopt a recurrent neural network with
parametric bias (RNNPB) with recurrent feedback from the output layer to the
input layer. As in \citep{tani2019accounting}, training and testing are decoupled and the
optimization is based on the back-propagation through time algorithm. The
optimization of the network parameters uses the prediction error between a
generated motor action and a reference action. Remarkably, this work analyses
the developmental dynamics of the parameter space in terms of prediction
error. Experiments are carried out on a simulated two degrees-of-freedom robot arm 
and on a Nao humanoid robot, where goal-directed actions are generated using the RNNPB.

An interesting series of studies has been produced by Lanillos and colleagues.
\citet{lanillos2018adaptive} present an architecture that combines generative models and a probabilistic framework inspired on some of the principles of predictive processing. The architecture is employed to estimate body configurations of a humanoid robot, using three modalities (proprioceptive, vision and touch). In the literature, the way how the brain integrates multi-modal streams in similar error minimization schemes is still under debate. Some authors suggest that the integration of different streams of unimodal sensory surprise occurs in hierarchically higher multimodal areas \citep{limanowski2013minimal,apps2014free,clark2013whatever,pezzulo2015active}, 
and therefore multimodal predictions and prediction errors would be generated \citep{friston2012prediction}. \cite{lanillos2018adaptive} apply
an additive formulation of unimodal prediction errors: (i) prior error, i.e. the "...error between the most plausible value of the body configuration and its prior belief"; (ii) proprioceptive error, i.e. the distance between joint angle readings and joint angle samples generated by a Normal distribution; (iii) visual error, i.e. the distance between observed end-effector image coordinates and those predicted by a visual generative model.

The proposed minimization scheme adjusts the prior on body configuration by summing up the additive multimodal error, while the system is exposed to multimodal observations.
As in Tani's work, training and testing are decoupled. 
The generative models are pre-trained using Gaussian Process Regression. In particular, a visual forward model maps 
proprioceptive data (position of three joints) to visual data (image coordinates of the end effector), whereas a 
proprioceptive model generates joint angles from a Normal distribution representing the joint states. Training data 
is recorded offline from a humanoid robot executing random trajectories. Another generative model is created for the 
tactile modality as a function of the visual generative model. This model is used in a second experiment to translate 
the end-effector positions to the spatial locations on the robot arm touched by an experimenter, in order to correct 
visual estimations.

A follow-up work \citep{oliver2019active} applies an active inference model for visuomotor coordination in the humanoid 
robot iCub. The framework controls two sub-systems of the robot body, i.e., the head and one arm.
An attractor model drives actions towards goals. Goals are specified in a visual domain -- encoded as linear 
velocity vectors towards a goal, whose 3D position is estimated using stereo vision and a color marker -- and 
transformed using a Moore-Penrose pseudoinverse Jacobian matrix into linear velocities in the 4D joint space of the robot.
Similarly, visual goals are transformed into joint velocity goals for the head sub-system. Authors assume 
normally distributed noise in the sensory inputs. Sensor variances and action gains are pre-tuned and fixed 
during the experiments.
Although no generative models are trained in this experiment (iCub's forward kinematics functions are used), 
authors show that minimizing Laplace-encoded free energy through gradient descent leads to reaching 
behaviours and visuo-motor coordination. Similarly, \cite{pezzato2020novel} present an active inference framework using a pre-coded controller and a generative 
function. The study aims at controlling the movements of an industrial robotic platform using active inference and at 
comparing its adaptivity and robustness to another state-of-the-art controller for robotic manipulators, namely the model reference adaptive controller (MRAC).

\cite{lanillos2020robot} extend the active inference implementation presented in \cite{oliver2019active}. In this study, 
the visual generative model is pre-trained using a probabilistic neural network (Mixture Density Network, MDN). Inverse
mapping is performed through the backward pass of the MDN of the most plausible Gaussian kernel. The system re-trains 
the network from scratch whenever the sensory inputs are too far from its predictions.
Differently from \citep{oliver2019active},
 visual inputs consist of movements estimated through an optical flow algorithm. The generative model thus maps joint angles to 
the 2D centroid of a moving blob detected from the camera. A deep learning classifier is then trained to label joint 
velocities and optical flow inputs as self-generated or not.

\cite{sancaktar2019end} apply a similar approach on the humanoid robot
Nao. The minimization scheme uses a pre-trained generative model for the
visual input, i.e., a convolutional decoder-like neural network. Training data
are collected through a combination of random babbling and kinesthetic
teaching. The generative model maps joint angles to visual inputs, as in
to \citet{lang2018deep}.  
When computing the likelihood for the gradient descent, the density defining
the visual input is created as a collection of independent Gaussian
distributions centered at each pixel. In the minimization scheme, the visual
prediction error multiplied by the inverse of the variance is calculated by
applying a forward pass and a backward pass to the convolutional decoder. The
approach is interesting, but studies have pointed at questionable
aspects about the biological plausibility of back-propagating errors. This refers, in particular, to the lack of
local error representations in ANNs and at the symmetry between forwards and
backwards weights, which is not always present in cortical networks
\citep{whittington2019theories}. 
As in the previous series of experiments, active inference is used to control
the robot arm movement in a reaching experiment. 

\cite{pio2016active} 
present a proof-of-concept implementation of a control scheme based on active inference using the 7 degrees-of-freedom arm of a simulated PR2 humanoid robot. 
The control scheme is adopted to perform trajectories towards predefined goals. Authors highlight that such a scheme eliminates the need of an inverse
model for motor control as ``action realizes the (sensory) consequences of
(prior) causes'' \cite[pp 9]{pio2016active}. A generative model maps causes to actions, where causes are seen as "forces that have some desired fixed point or orbit"\cite[pp 9]{pio2016active}, as sensed by proprioception. Proprioceptive predictions 
are thus realized in an open-loop fashion, by means of reflex arcs.

This framework -- which employs a hierarchical generative model -- minimizes the KL-divergence 
between the distribution of the agent's priors and that of the true posterior distribution, which represents the updated belief given the evidence. 
Authors point out that more complex behaviours require the design of equations of motion. The question on the scalability of such an approach for cognitive robotics remains open.

Although not adopting an active inference approach, \cite{schillaci2020}
present a study where intrinsically motivated behaviors are driven by error
minimization schemes in a simulated robot.  
The proposed architecture generates exploratory behaviors towards
self-generated goals, leverages computational resources and regulates goal selection and the balance between
exploitation and exploration through a multi-level monitoring of prediction error dynamics.  
The work is framed within the study of the underlying mechanisms of motivation
and the emergence of emotions that drive behaviors and goal selection to
promote learning. Scholars such as \cite{van2017affective}, \cite{kiverstein2019feeling} and \cite{hsee1991velocity} argue 
that what motivates engagement in a behavior is not just the final outcome, but the satisfaction that emerges from the pattern 
and the velocity of an outcome over time\footnote{Here we intend the desired outcome of an event or of an activity. As for the velocity of an outcome, we intend the velocity, or the rate, at which such desired goal is achieved. In the context of learning, a goal could be merely the reduction of prediction error. The velocity of the outcome here would correspond to the rate of reduction of the prediction error, i.e., how fast or slow is prediction error minimised.}. ``If one [...] assumes that people not only passively
experience satisfaction, but actively seek satisfaction, then one
can infer an interesting corollary from the velocity relation:
People engage in a behavior not just to seek its actual outcome,
but to seek a positive velocity of outcomes that the behavior
creates over time'' \citep[pp, 346]{hsee1991velocity}.

The system proposed by \cite{schillaci2020} monitors prediction error dynamics over time and at different levels, driving behaviours 
towards those goals that are associated to specific patterns of prediction error dynamics. The system also 
modulates exploration noise and leverages computational
resources according to the dynamics of the overall learning
performance. Learning is performed in an online fashion, where image features
-- compressed using a pre-trained convolutional autoencoder -- are fed into a
self-organizing neural network for unsupervised goal generation and into an
inverse-forward models pair for movement generation and prediction error
monitoring. The models are updated in an online fashion and an episodic
memory system is adopted to reduce catastrophic forgetting issues. 
Actions are generated towards goals associated with the steepest descent in low-level prediction error dynamics. 

A similar approach for the self-generation of goals has been employed by \cite{annabi2020autonomous} in a simulated experiment 
where a two degrees-of-freedom robotic arm has to learn how to write digits. The proposed architecture learns sequences of motor primitives based on a free energy minimization approach. 
The system combines recurrent neural networks for trajectories encoding with a self-organising system for goal estimation, which is trained on 
data generated through random behaviours. In the experiments, the system incrementally learns motor 
primitives and policies, using a predefined generative forward model. Free energy minimization is used for action selection.

\cite{zhong2018afa} present a hierarchical model consisting of a series of
repeated stacked modules to implement active inference in simulated agents. 
Each layer of the network contains different modules, including generative
units implemented as convolutional recurrent networks (Long Short-Term Memory networks, LSTM). In the hierarchical architecture,
predictions and prediction errors flow in top-down and bottom-up directions,
respectively. Generative units are trained in an off-line learning session during two
simulated experiments. 

It is worth noting that all the works reviewed in this section make use of different forms of prediction error minimization schemes to obtain working models and controllers.

\subsection{Non-robotic implementations}
A wide amount of non-robotic studies on predictive processing have been produced during the last years. 
This section opens only a small window on this
literature. Nevertheless, 
 promising directions for cognitive robotics research on predictive processing can be characterised from the few samples reported here.

The issue of scalability highlighted on the active inference study of \cite{pio2016active} 
is also apparent in the work of \cite{baltieri2019active}, where the authors design an active inference based linear quadratic Gaussian controller to 
manipulate a one degree-of-freedom system. The study aims at showing that such a controller can achieve goal positions without the need of an efference copy, 
as in optimal control theory (OCT). 

Similar basic proofs-of-concept are presented by
\cite{tschantz2020learning} and \cite{baltieri2017active}, where active inference is used to model bacterial chemo-taxis in a minimal simulated agent. \cite{tschantz2020learning} focus on an action-oriented model that employ goal-directed (instrumental) and information-seeking (epistemic) behaviors when learning a generative model. Different error minimization strategies are tested, generating epistemic, instrumental, random behaviours or expected free energy driven ones.
%Expected free-energy has a direct effect on the learning of the models, when epistemic models are learned, these are better for performing chemo-taxis. In their view, ``
Authors show that active inference balances exploration and exploitation and
suggest that ``[they] are both complementary perspectives of the same objective
function -- the minimization of expected free energy.''
\cite[pp.19]{tschantz2020learning}. 
The model is not hierarchical, but it fully exploits the
proposals of active inference. In the other interesting proof-of-concept, \cite{baltieri2017active} present a Braitenberg-like vehicle where behaviors are modulated according to predefined precision weights.  

\cite{friston2015active} also addresses the exploration-exploitation dilemma. Authors
argue that, when adopting Bayes optimal behavior under the free energy principle,
epistemic, intrinsic value is maximized until there is no further information
gain, after which exploitation is assured through 
maximization of extrinsic value, i.e., the utility of the result of an action.
In fact, epistemic actions can bring the agent far from a
goal. Nonetheless, they can be used to plan a path to a goal with greater confidence.
Adopting the formalism of partially observed
Markov decision processes, authors present a simulated experiment where an agent,
i.e., a rat, navigates through a T-shaped maze, to show the role of   epistemic value in resolving uncertainty about goal-directed behavior. 
Moreover, the authors discuss an aspect of the Bayesian framework, that is,
the role of the precision -- i.e., the inverse of the variance -- of the posterior belief -- which is estimated from the prior belief and the likelihood of the evidence --  about control states\footnote{In the generative model, a control state corresponds to the hidden cause of an action. ``This means the agent has to infer its behavior by forming beliefs about control states, based upon the observed consequences of its action''
\citep[pp. 190]{friston2015active}.} as a
message passing channel. Under this view, precision is associated with
dopaminergic responses, which has been interpreted in terms of changes in
expected value (e.g. reward prediction errors). In brief, changes in precision
would correlate with changes in exploratory or exploitative behaviors. 

In a follow-up study, \cite{schwartenbeck2019computational} present an architecture that has an 
implicit weighting of the exploitation and (goal-directed) exploration tendencies, 
determined by the precision of prior beliefs and the degree of uncertainty about the world. 
Two mechanisms for goal-directed exploration are implemented in the rat-within-a-maze simulated setup: 
 \textit{model parameter} exploration and \textit{hidden state} exploration. In the former active learning strategy, the agents
forage for information about the correct parameterization of the observation model, in the 
study represented as a Markovian model. Here, parameters are the set of arrays encoding the Markovian 
transition probabilities, i.e., the mapping between hidden states and observations and the transition between hidden states.
In the latter active inference strategy, 
agents aim at gathering information about the current (hidden) state of the world, for example the current context. In particular, they
sample the outcomes that are associated with a high uncertainty,
only when these are informative for the representation of the task structure. 
Similarly to a standard intrinsic motivation approach, authors appeal to the need 
of random sampling when the uncertainty about model parameters and hidden states 
(goal-exploration strategies) fails to inform behavior. The aim of this work is to understand 
``the generative mechanisms that underlie information gain and its trade-off with reward maximization'' 
\cite[pp. 45]{schwartenbeck2019computational}, but, as authors notice, how to scale up these 
mechanisms to more complicated tasks is still an open challenge.

Precision weighting is also one of the main focuses of the predictive coding study carried out by
\cite{oliva2019development}. Interestingly, the authors analyze the variations
of the precision of prior prediction of a recurrent (S-CTRNN) generative model
over a developmental process. The model learns to estimate stochastic time
series (two-dimensional trajectory drawings), thus providing an estimate of
the variance of the input data. The framework ``shares crucial properties with
the developmental process of humans in that it naturally switches from a
strong reliance on sensory input at an early learning stage to a proper
integration of sensory input and 
own predictions at later learning stages''
\cite[pp. 254]{oliva2019development}. This is correlated to a reduction of the
prediction error and the estimated (prior) variance over time during
learning. Some formulations of the problem in this work are, however, problematic, 
çin particular, in \citep{oliva2019development} the posterior is 
computed naively by multiplying the likelihood and the prior using the 
basic Bayesian formula, and learning is performed only for 
maximizing the likelihood.
In a follow-up work
\citep{philippsen2019predictive}, the framework is applied to simulate the
generation of representational drawings -- i.e., drawings that represent
objects -- in infants and chimpanzees. Authors observe that stronger reliance
on the prior (hyper-prior) enables the network to perform representational
drawings as those produced by children, whereas a weak reliance on the prior produces
highly accurate lines but fails to produce missing parts of the
representational drawings, as observed in chimpanzees. Results suggest that chimpanzees'
and humans'  
``differences in representational drawing behavior might be explainable by the degree to which they take prior information into account'' 
\cite[pp. 176]{philippsen2019predictive}. 

\cite{allen2019body} study active inference in a multimodal domain, simulating
interactions between interoceptive cardiac cycle and exteroceptive (visual)
perception.
The work hypothesizes that effects of cardiac timing on perception could arise as a function of periodic sensory attenuation. 
This study does not involve any robotic implementation nor any learning or control task. However, related 
implementations are mostly missing in the literature, therefore we believe it is worth being mentioned in this review.   

\section{Discussion}
\label{sec:dis}

This work has reviewed a series of robotics and non-robotics studies that have adopted the paradigm of predictive processing under different forms. Tables
\ref{tab:pp_checkmarks} and \ref{tab:training} provided a general overview of
the main aspects as well as the differences of these studies. 

It is certainly standing out to which length the robotics research and the non-robotics models 
have addressed tasks that go beyond perception and
motor control, which have been traditionally the focus of predictive processing
studies. 
Limited cognitive robotics research has addressed the scaling up of the predictive 
processing paradigm towards higher cognitive capabilities. 
Computational studies on minimal simulated systems
have suggested that specific aspects, such as precision weighting, may bridge this gap.  

Embodied robotic systems seem to be the most appropriate experimental platforms
not only for studying cognitive development within the predictive processing framework, but
also for extending this framework to a broader range of modalities and
behavioral possibilities. In fact, another aspect of the robotics researches reviewed in this paper worth
highlighting, is that almost the totality of
them\footnote{\cite{lanillos2018adaptive} address also the tactile modality in
  their study, but do not fully integrate it in the error minimization scheme.}
address only proprioception and a single exteroceptive modality,
i.e., vision. 
Little attention in the
robotics community has been posed on how multiple exteroceptive modalities -- for example, vision, haptic, auditory, etc. --, as well as interoceptive ones \citep{seth2018being}, can be integrated in
prediction error minimization schemes. 
Studies such as those from 
\citet{tschantz2020learning}, \citet{friston2015active}, \citet{schwartenbeck2019computational} and \citet{schillaci2020}
have discussed epistemic and emotional value, homeostatic drives and
intrinsic motivation that regulate behaviors. 
Interesting research directions for robotics should include extending this to multimodal
self-generated goals and to combinations of fixed homeostatic goals and dynamic
ones.

Another important point concerns precision weighting, as in predictive processing 
this is assigned a prominent role in behavior and goal
regulation, as well as in perceptual optimization processes. 
Further cognitive robotics study should explore this path.  
Most of the non-robotic implementations adopt a Bayesian or probabilistic formalization of error
minimization schemes. This allows an elegant formulation of the precision in weighting schemes, which consists of the inverse of the variance of the
prior and posterior distributions. However, alternative strategies are available for implementing precision weighting-like processes in non-probabilistic
models, including the modulation of neuronal activation or of synaptic weights in artificial neural networks, 
modulation of firing rates in spiking neural networks, dopaminergic modulation, and the like. There is a wide literature on sensor fusion techniques in the machine learning community which focuses on very related challenges, such as the learning and modulation of the relevance of single sensors in multimodal and predictive settings \citep{fayyad2020deep}.

A common denominator in all the reviewed implementations is the use of
predictions for guiding behavior. 
However, the implementations adopt different machine learning tools. 
Works that follow strictly the active inference principles make use of Bayes as their main 
tool. 
It is still an open question how all other approaches should be
considered in the wider predictive processing framework.  
So far, most robotics implementations make use of non-variational deep networks as their main tool. 
However, the bias of using the Bayesian framework, in non-robotics implementations,
might hinder the search for other approaches that could have advantages,  
importantly, in terms of computational cost and the complexity of designing
generative models to produce coherent and scaled-up behaviors. 

Predictive processing emphasizes the prediction-based learning of a generative
model, which predicts incoming sensory signals \citep{clark2015embodied}.
In optimal control theory, a high computational complexity is required for
learning to predict sensory consequences by means of the efference copy
and the inverse model. In predictive processing accounts, this complexity
is mapped to the learning of a generative model
during hierarchical perceptual and active inference
(e.g. \cite{friston2011optimal,friston2012active}).
In this regard, it is still unclear how generative models should be learned, due
to the complexity that implies modeling the richness of the entire environment
\citep{tschantz2020learning}.
Action-oriented models are a common approach to solve this issue by
learning and generating inferences that allow adaptive behavior, even when the 
world is not modelled in a precise manner
(e.g. \cite{tschantz2020learning,baltieri2017active,pezzulo2017model}).
It is worth highlighting that despite the relevance of learning for belief
updating, most non-robotic computational
work focuses on inference and not on learning. Actually, learning is almost
absent here.

The few non-robotic models that focus on 
learning of generative models are based on the expected free energy
formulations and use very simplified agents and behaviors
\citep{tschantz2020learning,baltieri2017active,ueltzhoffer2018deep,millidge2020deep}.
On the contrary, some cognitive robotics implementations do have 
the emphasis slightly shifted towards the learning of generative models
(e.g. \cite{lanillos2020robot,ahmadi2017can,idei2018neurorobotics,schillaci2020,schillaci2020intrinsic}). 
Yet, learning and testing are decoupled in many of these studies and, in 
particular, in those adopting probabilistic methods. This is likely due to the 
challenges in implementing online learning of probabilistic models, especially in the context of high-dimensional sensory and motor spaces.

It is worth pointing out that, 
in cognitive robotics, a variety of learning methods are used and just few
of these are equivalent to the free energy principle formulations. Nonetheless,
agents and behaviors used are much more complex. 
For cognitive robotics, it is very relevant to explore the reach and possibilities of using
generative models for perception, action, and planning.
More importantly, there is a special interest on the tools and methods that can be used for the learning of
these models, an area that has been unattended in non-robotic models using
predictive processing principles.

Finally, limited attention has been posed on the temporal aspect of prediction error dynamics
 \citep{kiverstein2019feeling,tschantz2020learning}. Prediction error patterns may be 
 associated with emotional experience \citep{joffily2013emotional}. In artificial systems, they are essential components for implementing
 intrinsically motivated exploration behaviors and artificial curiosity
\citep{oudeyer2007intrinsic,schillaci2020intrinsic,baldassarre2013intrinsically,graziano2011artificial}. Recent studies 
suggest that error dynamics may influence the regulation of
computational resources \citep{schillaci2020} and the emotional valence of
actions \citep{joffily2013emotional}. 
We believe that prediction error dynamics represent a promising tool in the exploration of more complex behaviours and
tasks in cognitive robotics under the predictive processing paradigm.

\subsection*{Acknowledgments}
Guido Schillaci has received funding from the European Union's Horizon 2020
research and innovation programme under the Marie Sklodowska-Curie grant
agreement No. 838861 (Predictive Robots). Predictive Robots is an associated
project of the Deutsche Forschungsgemeinschaft (DFG, German Research
Foundation) Priority Programme "The Active Self". 

Verena Hafner has received funding from the Deutsche Forschungsgemeinschaft (DFG, German Research
Foundation) Priority Programme "The Active Self" - 402790442 (Prerequisites for the Development of an Artificial Self).

Bruno Lara and Alejandra Ciria have received funding from the Alexander von Humboldt 
Foundation from the project "Predictive Autonomous Behaviour Internal Models and Predictive Self-regulation".

The authors would like to thank the anonymous reviewer for
his/her thorough reading of our manuscript. His/her comments 
helped greatly to improve the first version submitted.

\bibliography{ppreview}  
\bibliographystyle{apalike}

\end{document}